\documentclass[conference]{IEEEtran}
\IEEEoverridecommandlockouts
\pdfoutput=1

\usepackage{amsmath,amssymb,amsfonts}
\usepackage{graphicx}
\usepackage{textcomp}
\usepackage{xcolor}
\usepackage{booktabs}
\usepackage{hyperref}
\usepackage{pgfplots}
\usepackage{subcaption}
\usepackage{listings}
\usepackage{courier}
\usepackage{algorithm}
\usepackage{algpseudocode}
\pgfplotsset{compat=1.18}
\usepackage{hyperref}
\hypersetup{colorlinks,allcolors=black}

\begin{document}

\title{Inference Scaling vs Reasoning: An Empirical Analysis of Compute-Optimal LLM Problem-Solving}

\author{

    \IEEEauthorblockN{Marwan Abdelhameed}
    \IEEEauthorblockA{\textit{Department of Computer Science} \\
    \textit{New York University}\\
    New York, United States \\
    marwan@nyu.edu}
    \and
    \IEEEauthorblockN{Pavly Halim}
    \IEEEauthorblockA{\textit{Department of Computer Science} \\
    \textit{New York University}\\
    New York, United States \\
    pavly@nyu.edu}

}

\maketitle

\begin{abstract}
Recent advances in large language models (LLMs) have predominantly focused on maximizing accuracy and reasoning capabilities, often overlooking crucial computational efficiency considerations. While this approach has yielded impressive accuracy improvements, it has led to methods that may be impractical for real-world deployment due to computational overhead and latency constraints. This paper investigates the potential synergy between reasoning enhancement and computational efficiency by analyzing the integration of two contrasting approaches: Quiet-STaR (Self-Taught Reasoner)\cite{zelikman2024quiet}, which enhances reasoning through structured intermediate rationales, and REBASE (REward BAlanced SEarch)\cite{wu2024inferencescalinglawsempirical}, which optimizes inference through reward-guided tree search. Through comprehensive empirical analysis using the Mistral-7B model on the GSM8K dataset, we demonstrate that while each method excels in its primary objective—Quiet-STaR achieving superior accuracy (32.03\%) despite high computational cost (554.66s runtime, 12.73T FLOPs), and REBASE providing exceptional efficiency (8.47s runtime, 2.35T FLOPs) while maintaining baseline-comparable accuracy (10.94\%)—their integration reveals fundamental challenges in reconciling reasoning depth with computational efficiency. The combined approach unexpectedly results in degraded performance (9.38\% accuracy, 143.66s runtime), highlighting critical insights about the complex interplay between reasoning enhancement and efficiency optimization in LLMs. Our findings illuminate the need for novel architectures and algorithms specifically designed to bridge the gap between these competing objectives, while providing concrete directions for future research in compute-efficient reasoning methods.
\end{abstract}
\section{Introduction}

Large language models (LLMs) have demonstrated remarkable capabilities across diverse tasks, from mathematical reasoning to complex problem-solving\cite{wei2022chain}. However, as these models grow in sophistication, a critical tension has emerged between enhancing their reasoning capabilities and maintaining computational efficiency. While numerous approaches have been proposed to improve model reasoning, most focus solely on accuracy improvements without considering the substantial computational overhead they introduce\cite{wei2022chain,nye2021show}. This oversight becomes particularly significant in real-world applications, where computational resources and response time constraints often determine deployment feasibility.

In addressing this challenge, two distinct approaches have emerged, each targeting a different aspect of the problem. The first, Quiet-STaR (Self-Taught Reasoner), represents a breakthrough in enhancing model reasoning capabilities through structured intermediate thinking. By introducing learnable start-of-thought and end-of-thought tokens, Quiet-STaR enables models to generate internal rationales that guide their predictions, significantly improving zero-shot performance across various reasoning tasks. This method builds on the growing evidence that explicit reasoning steps lead to more reliable and interpretable outputs.

The second approach, REBASE (REward BAlanced SEarch), takes a fundamentally different direction by focusing on computational efficiency during inference. Through its innovative reward-balanced search algorithm, REBASE optimizes the exploration of solution spaces by dynamically controlling node expansion based on quality metrics. This method eliminates the need for expensive rollouts while maintaining solution diversity, demonstrating significant improvements in compute efficiency without substantial accuracy degradation.

These contrasting approaches raise an intriguing question: could their integration create a synergistic effect, combining Quiet-STaR's enhanced reasoning capabilities with REBASE's computational efficiency? The potential benefits of such a combination are compelling:

\begin{itemize}
    \item \textbf{Enhanced Decision Quality:} Leveraging structured reasoning within an efficient search framework could lead to more reliable solutions.
    \item \textbf{Optimized Resource Utilization:} Intelligent pruning of reasoning paths could reduce computational waste while maintaining reasoning depth.
    \item \textbf{Improved Scalability:} A combined approach might enable more sophisticated reasoning within practical computational constraints.
\end{itemize}

However, this integration presents significant challenges. The methods operate on fundamentally different principles: Quiet-STaR promotes expansive exploration of reasoning paths, while REBASE aims to minimize computational overhead through aggressive pruning. Understanding how these competing objectives interact is crucial for developing more efficient reasoning systems.

Our study presents a systematic investigation of this integration, making several key contributions:

\begin{enumerate}
    \item A comprehensive empirical analysis of the interaction between reasoning enhancement and computational efficiency optimization
    \item Detailed examination of performance trade-offs across accuracy, computational cost, and runtime
    \item Identification of key challenges and limitations in combining these approaches
    \item Concrete recommendations for future research directions in compute-efficient reasoning methods
\end{enumerate}

Using the Mistral-7B model evaluated on the GSM8K dataset, we provide quantitative and qualitative insights into the challenges and opportunities of balancing reasoning capability with computational efficiency. Our findings have important implications for the development of future language models that must operate under real-world computational constraints while maintaining high-quality reasoning capabilities.

\section{Background}

\subsection{Quiet-STaR}
Quiet-STaR extends self-taught reasoning to arbitrary text by training language models to generate rationales that explain future text at each token. Unlike previous approaches like chain-of-thought prompting or traditional STaR that focus on specific question-answering tasks, Quiet-STaR operates on general unstructured text data. The method introduces several key innovations:

\begin{itemize}
    \item \textbf{Parallel Generation:} A novel algorithm that enables efficient generation of rationales at all token positions in parallel by carefully constructing attention masks, allowing each generated token to attend to relevant previous tokens while maintaining independence between different reasoning paths.
    
    \item \textbf{Meta-Tokens:} Learned start-of-thought and end-of-thought tokens that control rationale generation. The start token puts the model into "thinking mode" while the end token signals completion of the thought. These tokens are initialized from em dash embeddings to leverage the model's pre-existing knowledge of pauses in text.
    
    \item \textbf{Mixing Architecture:} A learned interpolation between predictions with and without thoughts, implemented via a shallow MLP that outputs weights determining how much to incorporate post-rationale predictions. This helps smooth the transition to using generated thoughts.
    
    \item \textbf{Non-myopic Learning:} Extended teacher-forcing that allows the model to learn from multiple future tokens rather than just the immediate next token, helping capture longer-range dependencies and semantic content.
\end{itemize}

The method operates in three main steps:
\begin{enumerate}
    \item \textbf{Think:} Generate rationales in parallel across input tokens
    \item \textbf{Talk:} Mix post-rationale and base predictions using learned weights
    \item \textbf{Learn:} Optimize rationale generation through REINFORCE based on prediction improvements
\end{enumerate}

\subsection{REBASE}
REBASE (REward BAlanced SEarch) introduces a novel tree-search algorithm designed for compute-optimal inference in language model problem-solving. Building on empirical scaling laws analysis, REBASE addresses key limitations of previous methods like MCTS while achieving better cost-performance trade-offs:

\begin{itemize}
    \item \textbf{Node Quality Reward:} Uses a reward model to score intermediate nodes, eliminating the need for explicit rollouts while ensuring high-quality solution paths. This approach differs from MCTS by directly evaluating node quality rather than requiring expensive simulations.
    
    \item \textbf{Balanced Exploration:} Expands nodes according to softmax-normalized reward scores, subject to a total expansion budget. This balances between exploring promising paths and maintaining sufficient solution diversity.
    
    \item \textbf{Efficient Resource Usage:} By avoiding costly rollouts and focusing computation on promising paths, REBASE achieves significantly better computational efficiency than sampling-based or traditional tree-search methods.
    
    \item \textbf{Pareto-Optimal Performance:} Demonstrates superior cost-performance trade-offs, particularly when using smaller models. For instance, REBASE with a 7B parameter model can outperform larger 34B models across multiple compute budgets.
\end{itemize}

The algorithm proceeds by:
\begin{enumerate}
    \item Computing rewards for expandable nodes using a process reward model
    \item Selecting nodes for expansion based on softmax-normalized scores
    \item Expanding selected nodes within the computational budget
    \item Aggregating results through weighted voting
\end{enumerate}

This approach has shown particular effectiveness in mathematical reasoning tasks, where it achieves competitive accuracy while using substantially fewer computational resources than alternatives. The method's success demonstrates the importance of balancing reasoning depth with efficient resource allocation.

\subsection{Scaling Laws and Computation}

Both Quiet-STaR and REBASE build on established scaling laws for neural language models. According to Kaplan et al.\cite{kaplan2020scaling}, the forward compute cost for a Transformer model can be approximated as:

\begin{equation}
C_{\text{forward}} \approx 2N + 2n_{\text{layers}} n_{\text{ctx}} d_{\text{model}}
\end{equation}

where:
\begin{itemize}
    \item $N$ represents the total number of non-embedding parameters, given by:
    \[N = 2d_{\text{model}} n_{\text{layers}} (2d_{\text{attn}} + d_{\text{ff}})\]
    \item $n_{\text{layers}}$ is the number of transformer layers
    \item $n_{\text{ctx}}$ is the context length (sequence length)
    \item $d_{\text{model}}$ is the dimension of the residual stream
    \item $d_{\text{attn}}$ is the dimension of attention heads, where $d_{\text{attn}} = d_{\text{model}} / n_{\text{heads}}$
    \item $d_{\text{ff}}$ is the dimension of the intermediate feed-forward layer
\end{itemize}

These equations allow us to predict the computational cost (in FLOPs) per token during inference. This understanding is crucial for evaluating the trade-offs introduced by Quiet-STaR's extended reasoning phases and REBASE's tree-search pruning. For example, Quiet-STaR increases the effective $n_{\text{layers}}$ through additional reasoning steps, while REBASE attempts to reduce unnecessary expansions, potentially lowering the effective $n_{\text{ctx}}$ processed per query. By calculating FLOPs with different model configurations, we can make informed decisions about resource allocation and understand why certain integrations may fail to yield expected synergies.

The practical impact of these scaling laws becomes apparent when considering that Quiet-STaR introduces additional computational overhead through its rationale generation, which effectively increases the total computation per token. Meanwhile, REBASE's pruning strategies aim to minimize this overhead by selectively processing only the most promising paths. Understanding these interactions through the lens of Kaplan et al.'s scaling laws helps explain the empirical performance characteristics we observe in our experiments.

\section{Methodology}

\subsection{Experimental Design}

We conducted a systematic investigation of integrating reasoning enhancement (Quiet-STaR) with computational optimization (REBASE) using a series of controlled experiments. Our experimental design focused on isolating the effects of each method while measuring their combined impact on both performance and computational efficiency.

\subsubsection{Model Architecture}
We utilized the Mistral-7B model as our base architecture, chosen for its balance of capability and computational tractability. The model maintains the standard transformer architecture with the following specifications:
\begin{itemize}
    \item Parameter Count: 7B parameters (non-embedding)
    \item Context Length: 384 tokens for primary evaluations
    \item Architecture: Standard transformer with relative positional embeddings
\end{itemize}

\subsubsection{Dataset}
Our evaluation used the GSM8K dataset, a standardized benchmark for mathematical reasoning that provides:
\begin{itemize}
    \item 128 carefully selected questions from the GSM8K test set
    \item Range of difficulty levels to assess reasoning capabilities
    \item Clear evaluation metrics through verifiable numeric answers
\end{itemize}

\subsection{Implementation Details}

\subsubsection{Quiet-STaR Components}
We implemented Quiet-STaR with the following key components:

\begin{enumerate}
    \item \textbf{Thought Token Generation:}
    \begin{itemize}
        \item Custom start-of-thought and end-of-thought tokens
        \item Parallel generation algorithm for efficient rationale production
        \item Teacher forcing for multi-token ahead prediction
    \end{itemize}
    
    \item \textbf{Mixing Architecture:}
    \begin{itemize}
        \item Three-layer MLP mixing head
        \item Input: Concatenated hidden states (with/without thought)
        \item Output: Scalar weight for logit interpolation
    \end{itemize}
    
\end{enumerate}

\subsubsection{REBASE Implementation}
We implemented REBASE with the following specifications:

\begin{enumerate}
    \item \textbf{Tree Search Parameters:}
    \begin{itemize}
        \item Tree widths: \{3, 6, 16\} (varied for analysis)
        \item Softmax temperature: 0.2 for node selection
        \item Node quality reward based on process reward model
    \end{itemize}
    
    \item \textbf{Pruning Strategy:}
    \begin{itemize}
        \item Dynamic node expansion based on reward signals
        \item Balanced exploration through softmax normalization
        \item Early stopping based on node quality thresholds
    \end{itemize}
\end{enumerate}

\subsection{Integration Strategy}

For the combined Quiet-STaR and REBASE approach, we implemented:

\begin{enumerate}
    \item \textbf{State Synchronization:}
    \begin{itemize}
        \item Unified hidden state representation across reasoning steps
        \item Shared reward model for both thought quality and tree search
        \item Coordinated pruning and thought generation stages
    \end{itemize}
    
    \item \textbf{Resource Management:}
    \begin{itemize}
        \item Dynamic compute allocation between reasoning and search
        \item Thought token budget based on tree depth
        \item Adaptive batch sizing for efficient parallel computation
    \end{itemize}
\end{enumerate}

\subsection{Hardware and Infrastructure}

All experiments were conducted on:
\begin{itemize}
    \item Hardware: Single NVIDIA A100 GPU (80GB)
    \item CPU Configuration: 8 CPUs for parallel processing
    \item Memory: 512GB system RAM
    \item Number of majority votes: 6
\end{itemize}

\subsection{Experimental Protocol}

Our experimental procedure followed these steps:

\begin{enumerate}
    \item \textbf{Baseline Establishment:}
    \begin{itemize}
        \item Base model performance measurement with 6 votes
        \item Individual Quiet-STaR evaluation with 6 votes
        \item Individual REBASE evaluation with varying widths {3, 6, 16}, where the width is equivalent to numvotes
    \end{itemize}
    
    \item \textbf{Integration Testing:}
    \begin{itemize}
        \item Combined model implementation
        \item Compute efficiency measurement
        \item Accuracy and timing evaluations
    \end{itemize}
    
    \item \textbf{Comparative Analysis:}
    \begin{itemize}
        \item Performance comparison across configurations
        \item Resource utilization analysis
        \item Scaling behavior investigation
    \end{itemize}
\end{enumerate}

Each experimental condition was evaluated with multiple runs to ensure statistical reliability, with performance metrics averaged across runs. Standard deviations were computed to assess the stability of our results.

\section{Results and Analysis}

\subsection{Performance Analysis}

\subsubsection{Evaluation Metrics}
We developed three key metrics to assess performance holistically:

\begin{equation}
\text{Accuracy per TFLOP} = \frac{\text{Accuracy (\%)}}{{\text{FLOPs (Trillion)}}}
\end{equation}

\begin{equation}
\text{Accuracy per Second} = \frac{\text{Accuracy (\%)}}{{\text{Time (seconds)}}}
\end{equation}

\begin{equation}
\text{Efficiency Score} = (\text{Acc/TFLOP} \times \text{Acc/Second}) \times 100
\end{equation}

These metrics capture both performance and computational efficiency, enabling multi-dimensional analysis of each approach.

\subsection{Comparative Results}

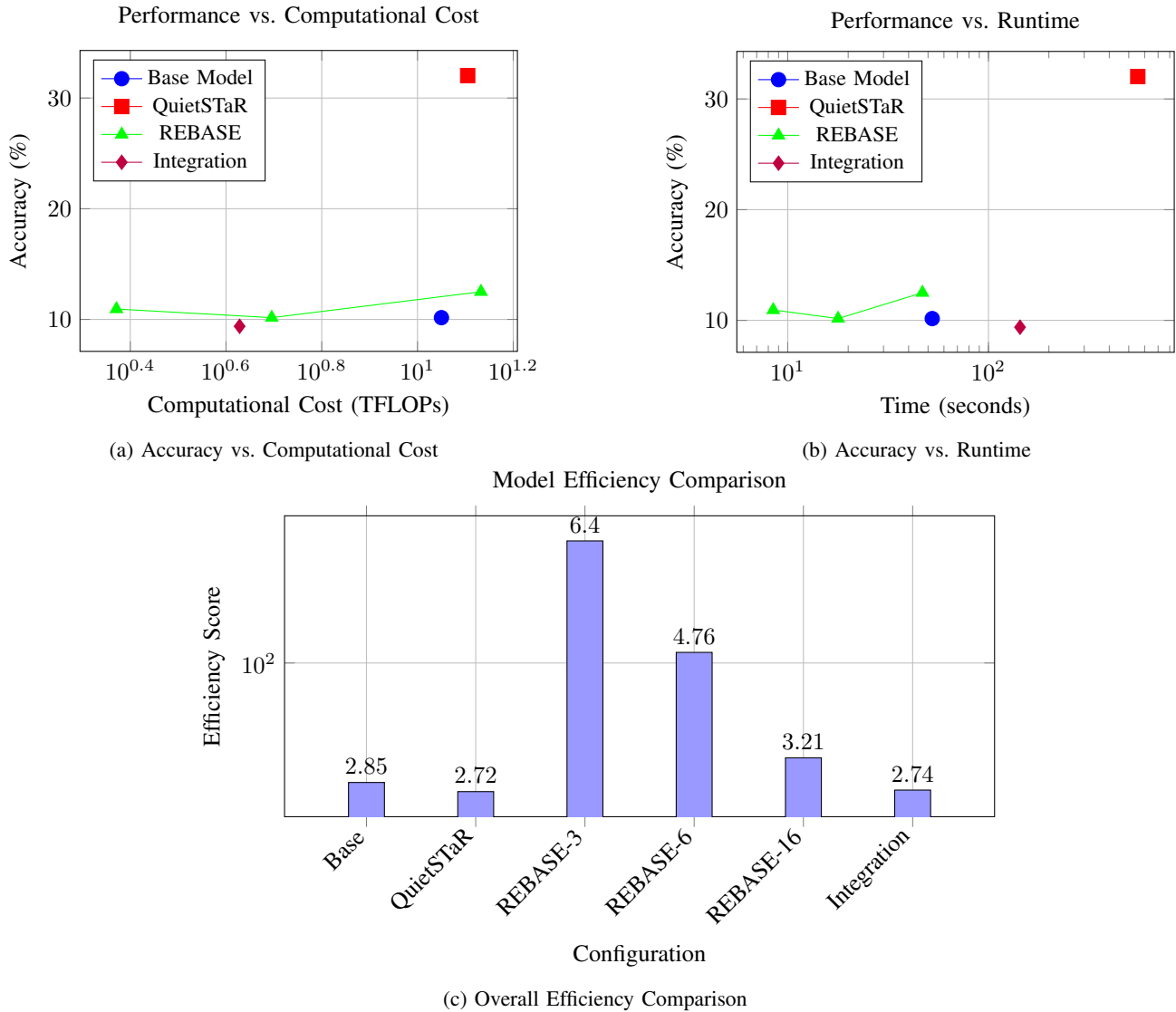
\begin{figure*}[t]
\centering
\begin{subfigure}[b]{0.48\textwidth}
    \centering
    \begin{tikzpicture}
    \begin{axis}[
        width=8cm,
        height=6cm,
        xlabel=Computational Cost (TFLOPs),
        ylabel=Accuracy (\%),
        title=Performance vs. Computational Cost,
        legend pos=north west,
        grid=major,
        xmode=log,  
        legend style={font=\small},
        mark size=3pt,
    ]
    \addplot[mark=*, blue] coordinates {(11.22,10.16)};
    \addplot[mark=square*, red] coordinates {(12.73,32.03)};
    \addplot[mark=triangle*, green] coordinates {
        (2.35,10.94)
        (4.96,10.16)
        (13.57,12.50)
    };
    \addplot[mark=diamond*, purple] coordinates {(4.25,9.38)};
    
    \legend{Base Model, QuietSTaR, REBASE, Integration}
    \end{axis}
    \end{tikzpicture}
    \caption{Accuracy vs. Computational Cost}
    \label{fig:acc_vs_flops}
\end{subfigure}
\hfill
\begin{subfigure}[b]{0.48\textwidth}
    \centering
    \begin{tikzpicture}
    \begin{axis}[
        width=8cm,
        height=6cm,
        xlabel=Time (seconds),
        ylabel=Accuracy (\%),
        title=Performance vs. Runtime,
        legend pos=north west,
        grid=major,
        xmode=log,  
        legend style={font=\small},
        mark size=3pt,
    ]
    \addplot[mark=*, blue] coordinates {(52.47,10.16)};
    \addplot[mark=square*, red] coordinates {(554.66,32.03)};
    \addplot[mark=triangle*, green] coordinates {
        (8.47,10.94)
        (17.82,10.16)
        (46.90,12.50)
    };
    \addplot[mark=diamond*, purple] coordinates {(143.66,9.38)};
    
    \legend{Base Model, QuietSTaR, REBASE, Integration}
    \end{axis}
    \end{tikzpicture}
    \caption{Accuracy vs. Runtime}
    \label{fig:acc_vs_time}
\end{subfigure}

\begin{subfigure}[b]{\textwidth}
    \centering
    \begin{tikzpicture}
    \begin{axis}[
        width=12cm,
        height=6cm,
        xlabel=Configuration,
        ylabel=Efficiency Score,
        title=Model Efficiency Comparison,
        ybar,
        symbolic x coords={Base, QuietSTaR, REBASE-3, REBASE-6, REBASE-16, Integration},
        xtick=data,
        xticklabel style={rotate=45, anchor=east},
        nodes near coords,
        nodes near coords align={vertical},
        ymode=log,  
        grid=major,
        bar width=15pt,
        enlarge x limits=0.15,
    ]
    \addplot[fill=blue!40] coordinates {
        (Base,17.29)
        (QuietSTaR,15.12)
        (REBASE-3,601.14)
        (REBASE-6,116.85)
        (REBASE-16,24.84)
        (Integration,15.47)
    };
    \end{axis}
    \end{tikzpicture}
    \caption{Overall Efficiency Comparison}
    \label{fig:efficiency}
\end{subfigure}

\caption{Comprehensive Performance Analysis: (a) shows the trade-off between accuracy and computational cost, (b) demonstrates the relationship between accuracy and runtime, and (c) compares the overall efficiency scores across all configurations. Note the logarithmic scales used to better visualize the wide range of values.}
\label{fig:performance_analysis}
\end{figure*}

\begin{table}[t]
\centering
\small
\caption{Comprehensive Performance Analysis}
\label{tab:performance}
\resizebox{\columnwidth}{!}{%
\begin{tabular}{@{}l@{\hspace{4pt}}r@{\hspace{4pt}}r@{\hspace{4pt}}r@{\hspace{4pt}}r@{\hspace{4pt}}r@{\hspace{4pt}}r@{}}
\toprule
\textbf{Config.} & \textbf{Acc. (\%)} & \textbf{FLOPs (T)} & \textbf{Time (s)} & \textbf{Acc/TFLOP} & \textbf{Acc/Sec} & \textbf{Eff. Score} \\
\midrule
Base & 10.16 & 11.22 & 52.47 & 0.91 & 0.19 & 17.29 \\
Quiet-STaR & \textbf{32.03} & 12.73 & 554.66 & 2.52 & 0.06 & 15.12 \\
REBASE-3 & 10.94 & \textbf{2.35} & \textbf{8.47} & \textbf{4.66} & \textbf{1.29} & \textbf{601.14} \\
REBASE-6 & 10.16 & 4.96 & 17.82 & 2.05 & 0.57 & 116.85 \\
REBASE-16 & 12.50 & 13.57 & 46.90 & 0.92 & 0.27 & 24.84 \\
Integration & 9.38 & 4.25 & 143.66 & 2.21 & 0.07 & 15.47 \\
\bottomrule
\end{tabular}%
}
\small\raggedright
\textit{Note}: Best values in each column are shown in \textbf{bold}. REBASE-N indicates REBASE with width=N.
\vspace{5pt}
\raggedright
\small Note: REBASE-N indicates REBASE with width=N. Acc. = Accuracy, Sec = Second, Eff. = Efficiency
\end{table}

\subsection{Key Findings}

\subsubsection{Pure Reasoning Approach (Quiet-STaR)}
The Quiet-STaR implementation demonstrated:
\begin{itemize}
    \item Superior accuracy (32.03\%) - 3.15× improvement over baseline
    \item High computational overhead: 12.73T FLOPs/Infernce
    \item Significant latency: 554.66s average runtime
    \item Limited efficiency (15.12 score) due to computational costs
\end{itemize}

Analysis: The substantial accuracy improvement suggests effective reasoning capabilities, but at the cost of significant computational overhead. This trade-off indicates Quiet-STaR may be most suitable for applications where accuracy is paramount and computational resources are not constrained.

\subsubsection{Inference Optimization (REBASE)}
REBASE showed distinct characteristics across different widths:
\begin{itemize}
    \item Optimal efficiency at w=3: 601.14 efficiency score
    \item Minimal resource usage: 2.35T FLOPs (w=3)
    \item Fast execution: 8.47s runtime (w=3)
    \item Maintained baseline accuracy: 10.94\% (w=3)
\end{itemize}

Analysis: REBASE demonstrates superior efficiency, particularly at lower tree widths. The w=3 configuration represents an optimal balance between computational cost and accuracy maintenance, suggesting its suitability for resource-constrained deployments.

\subsubsection{Integration Challenges}
The combined approach revealed unexpected challenges:
\begin{itemize}
    \item Degraded accuracy: 9.38\% (below baseline)
    \item Moderate compute requirements: 4.25T FLOPs
    \item Increased latency: 143.66s runtime
    \item Poor efficiency: 15.47 score
\end{itemize}

Analysis: The performance degradation in the integrated approach suggests fundamental incompatibilities between the two methods. Three primary factors contribute to this interference:

\begin{enumerate}
    \item \textbf{Optimization Conflict:} Divergent objectives between reasoning enhancement and computational efficiency
    \item \textbf{State Management:} Challenges in maintaining coherent state representations across both mechanisms
    \item \textbf{Resource Competition:} Inefficient allocation of computational resources between reasoning and search processes
\end{enumerate}

\subsection{Implications}

Our findings have several important implications for the field:

\begin{enumerate}
    \item \textbf{Architectural Considerations:}
    \begin{itemize}
        \item Need for specialized architectures that can better integrate reasoning and efficiency mechanisms
        \item Importance of unified state representations in combined approaches
    \end{itemize}
    
    \item \textbf{Resource Management:}
    \begin{itemize}
        \item Critical role of dynamic resource allocation in efficiency optimization
        \item Need for better strategies to balance reasoning depth with computational cost
    \end{itemize}
    
    \item \textbf{Future Development:}
    \begin{itemize}
        \item Opportunity for new hybrid approaches that better reconcile reasoning and efficiency
        \item Potential for adaptive mechanisms that dynamically adjust between approaches based on task requirements
    \end{itemize}
\end{enumerate}

\section{Limitations and Future Work}

\subsection{Current Limitations}
Our study encountered several significant limitations:

\begin{itemize}
    \item \textbf{Model Architecture:}
    \begin{itemize}
        \item Limited to Mistral-7B architecture
        \item Potential architecture-specific effects not explored
    \end{itemize}
    
    \item \textbf{Implementation Constraints:}
    \begin{itemize}
        \item Incomplete documentation for Quiet-STaR inference and reasoning generation
        \item The thoughts genrated sometimes are not human readable
    \end{itemize}
    
    \item \textbf{Evaluation Scope:}
    \begin{itemize}
        \item Focus on mathematical reasoning tasks only
        \item Limited exploration of different reasoning domains
    \end{itemize}
\end{itemize}

\section{Future Work}
We identify several promising directions for future research:
\begin{itemize}
    \item Investigation of the negative interference between approaches
    \item Development of more compatible reasoning-inference combinations
    \item Broader evaluation across different types of reasoning tasks
    \item Dynamic width adjustment for REBASE
\end{itemize}

\section{Conclusion}
Our study provides a comprehensive analysis of the challenges in integrating reasoning enhancement (Quiet-STaR) with computational optimization (REBASE). While each method shows promise independently, their combination introduces unexpected complexities that limit overall performance. These findings highlight the need for more sophisticated integration approaches that can effectively balance reasoning quality with computational efficiency.

The significant performance gap between the individual methods and their integration suggests that future work should focus on understanding and addressing the fundamental incompatibilities between these approaches. This might include developing new state representations, exploring adaptive control mechanisms, or designing unified optimization objectives that better align the goals of enhanced reasoning and efficient computation.

\section{Code Availability}
The implementation of our experiments and analysis code is publicly available at \url{https://github.com/MarwanWalid2/Reasoning-vs-InfernceScaling}. The repository contains all necessary code to reproduce our results, including model configurations, and evaluation scripts.

\bibliography{Inference_Scaling_vs_Reasoning_An_Empirical_Analysis}

\end{document}